\documentclass[sigconf]{acmart}

\usepackage{graphicx}
\usepackage{enumitem}
\usepackage{float}
\usepackage{wrapfig,lipsum,booktabs}
\usepackage{multirow}
\usepackage{amsmath}
\usepackage{amsthm}
\usepackage{balance}
\newtheorem{theorem}{Theorem}

\newtheorem{lemma}{Lemma}
\newtheorem{prop}{Proposition}

\usepackage{hyperref}
\usepackage{url}
\usepackage{titlesec}
\usepackage{xcolor}

\AtBeginDocument{%
  \providecommand\BibTeX{{%
    \normalfont B\kern-0.5em{\scshape i\kern-0.25em b}\kern-0.8em\TeX}}}

\copyrightyear{2023}
\acmYear{2023}
\setcopyright{acmlicensed}
\acmConference[KDD '23] {Proceedings of the 29th ACM SIGKDD Conference on Knowledge Discovery and Data Mining}{August 6--10, 2023}{Long Beach, CA, USA.}
\acmBooktitle{Proceedings of the 29th ACM SIGKDD Conference on Knowledge Discovery and Data Mining (KDD '23), August 6--10, 2023, Long Beach, CA, USA}
\acmPrice{15.00}
\acmISBN{979-8-4007-0103-0/23/08}
\acmDOI{10.1145/3580305.3599390}

\titleformat{\section}[block]{\normalfont\Large\bfseries}{\thesection}{1em}{}
\titlespacing*{\section}{0pt}{*0.6}{*0.6}

\begin{document}

\title{Improving the Expressiveness of $K$-hop Message-Passing GNNs by Injecting Contextualized Substructure Information}

\author{Tianjun Yao}
\email{tianjun.yao@mbzuai.ac.ae}
\orcid{0009-0006-0553-2809}
\affiliation{%
  \institution{Mohamed bin Zayed University of Artificial Intelligence}
  \city{Abu Dhabi}
  \country{UAE}
}

\author{Yingxu Wang}
\email{yingxv.wang@gmail.com}
\orcid{0000-0003-3284-1464}
\affiliation{%
  \institution{Mohamed bin Zayed University of Artificial Intelligence}
  \city{Abu Dhabi}
  \country{UAE}
}

\author{Kun Zhang}
\email{kunz1@cmu.edu}
\orcid{0000-0002-0738-9958}
\affiliation{%
  \institution{Carnegie Mellon University \& Mohamed bin Zayed University of Artificial Intelligence}
  \city{Pittsburgh}
  \state{PA}
  \country{USA}
}


\author{Shangsong Liang}
\email{liangshangsong@gmail.com}
\orcid{0000-0003-1625-2168}
\affiliation{%
  \institution{ Mohamed bin Zayed University of Artificial Intelligence}
  \city{Abu Dhabi}
  \country{UAE}
}


\renewcommand{\shortauthors}{Tianjun Yao, et al.}


\begin{abstract}
    Graph neural networks (GNNs) have become the \textit{de facto} standard for representational learning in graphs, and have achieved state-of-the-art performance in many graph-related tasks; however, it has been shown that the expressive power of standard GNNs are equivalent maximally to 1-dimensional Weisfeiler-Lehman (1-WL) Test. Recently, there is a line of works aiming to enhance the expressive power of graph neural networks. One line of such works aim at developing $K$-hop message-passing GNNs where node representation is updated by aggregating information from not only direct neighbors but all neighbors within $K$-hop of the node. Another line of works leverages subgraph information to enhance the expressive power which is proven to be strictly more powerful than 1-WL test. In this work, we discuss the limitation of $K$-hop message-passing GNNs and propose \textit{substructure encoding function} to uplift the expressive power of any $K$-hop message-passing GNN. We further inject contextualized substructure information to enhance the expressiveness of $K$-hop message-passing GNNs. Our method is provably more powerful than previous works on $K$-hop graph neural networks and 1-WL subgraph GNNs, which is a specific type of subgraph based GNN models, and not less powerful than 3-WL. Empirically, our proposed method set new state-of-the-art performance or achieves comparable performance for a variety of datasets. Our code is available at \url{https://github.com/tianyao-aka/Expresive_K_hop_GNNs}.
\end{abstract}





\maketitle

\section{Introduction}
\label{intro}

Graph-structured data is ubiquitous in many real-world applications ranging from social network analysis~\cite{fan2019graph}, drug discovery~\cite{jiang2020drug}, personalized recommendation~\cite{he2020lightgcn} and bioinformatics~\cite{gasteiger2021gemnet}. In recent years, Graph Neural Networks (GNNs) have seized increasing attention due to their powerful expressiveness and have become dominant approaches for graph-related tasks. 
message-passing Graph Neural Networks (MPGNNs) are the most common types of GNNs, thanks to their efficiency and expressivity. MPGNNs can be viewed as a neural version of the
1-Weisfeiler-Lehman (1-WL) algorithm~\cite{weisfeiler1968reduction}, where colors are replaced by continuous feature vectors and neural networks are used to aggregate over node neighborhoods~\cite{morris2019weisfeiler}. By iteratively aggregating neighboring node features to the center node, MPGNNs learn node representations that encode their local structures and feature information. A graph readout function can be further leveraged to pool a whole-graph representation for downstream tasks such as graph classification. 

Despite the success of MPGNNs, it is proved in recent developments that the expressive power of MPGNNs is bounded by 1-WL isomorphism test~\citep{morris2019weisfeiler,xu2018powerful}, i.e, standard MPGNNs or 1-WL GNNs cannot distinguish any (sub-)graph structure that 1-WL cannot distinguish; for instance, for any two $n$-node $r$-regular graphs, standard MPGNNs will output the same node representations.

\begin{figure*}[h]
\vspace{-20pt}
\centering
\includegraphics[width=0.7\textwidth]{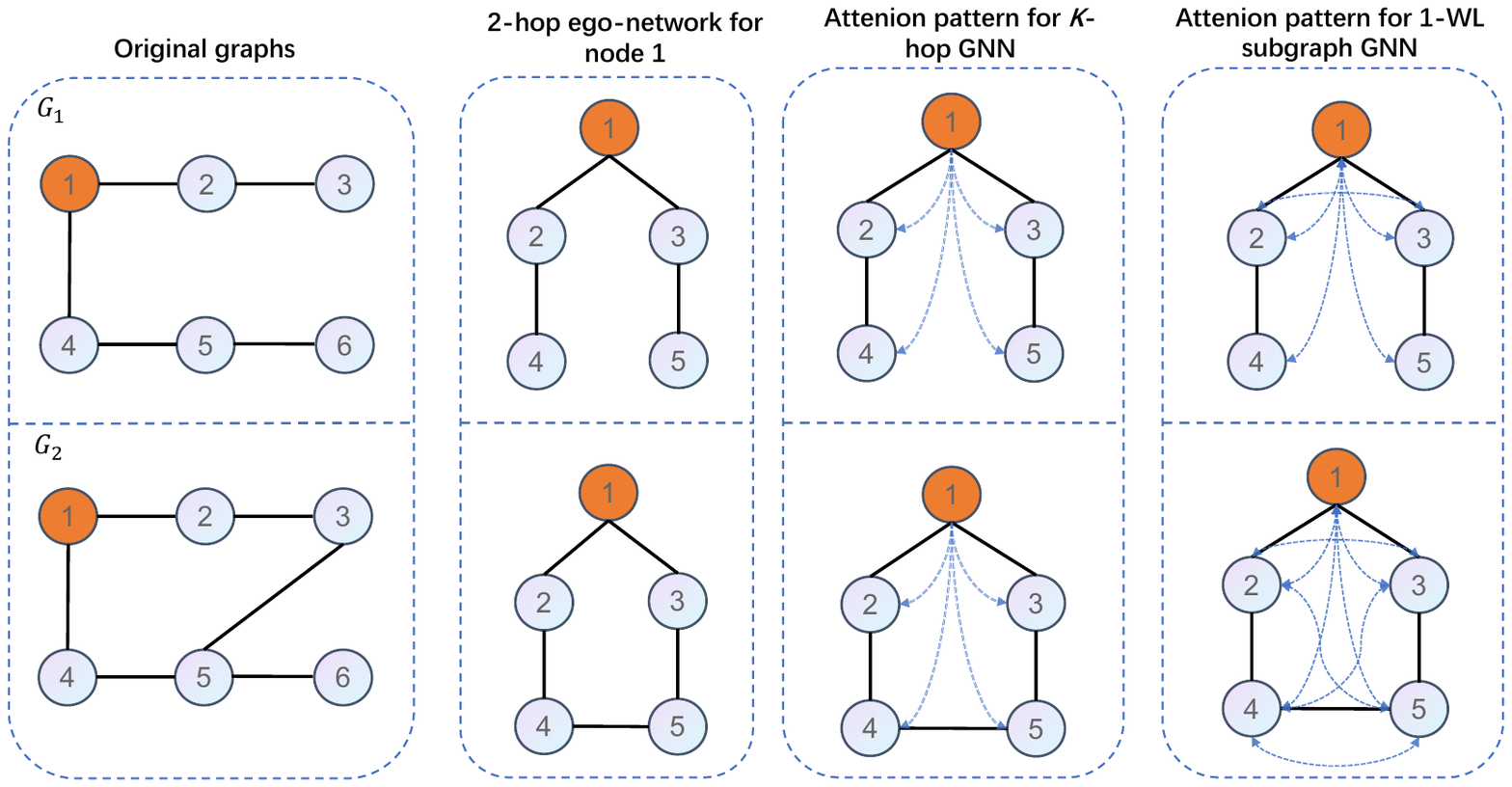}
\vspace{-40pt}
\caption{One example where node 1 in $G_1$ and $G_2$ induces the same attention pattern given a 1-layer 2-hop message-passing GNN method, and induces different attention patterns using a 1-layer 1-WL subgraph GNN where a 2-layer base GNN encoder is used. }
\label{fig:motivation}
\end{figure*}

Since then, a few works have been proposed to enhance the expressivity of MPGNNs. Methods proposed in ~\citep{morris2019weisfeiler,chen2019equivalence,maron2019provably,morris2021weisfeiler,maron2019universality,geerts2022expressiveness} aimed at approximating high-dimensional WL tests. However, these methods are computationally expensive and not able to scale well to large-scale graphs. In light of this, some other recent works sought to develop new GNN architectures with improved expressiveness while still being efficient in terms of time and space complexity. One line of works aimed to achieve this by leveraging subgraph information\citep{bevilacqua2021equivariant,maron2020learning,cotta2021reconstruction,zhang2021nested,zhao2021stars}, among which Nested GNN\cite{zhang2021nested} and GIN-AK\cite{zhao2021stars} developed GNN models that hash neighboring subgraphs instead of direct neighbors, effectively representing nodes by means of GNNs applied to their enclosing ego-networks followed by a graph readout function $\mathcal{R}$, which we term as \textit{1-WL subgraph GNN}. 

Another line of works leverages the notion of $K$-hop message passing\citep{abu2019mixhop,wang2020multi,chien2020adaptive,brossard2020graph,nikolentzos2020k}, where node representation is updated by iteratively aggregating information from not only direct neighbors but also neighbors within $K$-hop of the node, which implicitly leverage subgraph information ($K$-hop ego-networks) to perform message passing. In this work, we establish the connection between $K$-hop message-passing GNNs and 1-WL subgraph GNNs by viewing their message-passing schemes as two different attention patterns. As shown in Figure \ref{fig:motivation}, $K$-hop message-passing GNN induces a uni-directional hub-spoke attention pattern, while 1-WL subgraph GNN induces a pairwise bidirectional attention pattern. Assuming node features for $G_1$ and $G_2$ are identical from a countable set, a 2-hop 1-layer GNN is not able to distinguish the structural disparity for the 2-hop ego-network induced by node 1 due to the same resulting attention patterns. However, a 1-layer 1-WL subgraph GNN with a 2-layer base GNN encoder is able to achieve it, thanks to its pairwise bidirectional attention pattern, assuming all functions in the GNN model are injective. Intuitively, the expressive power of 1-WL subgraph GNN is stronger than $K$-hop message-passing GNNs with the same receptive field as 1-WL subgraph GNN is able to learn representation that reflects the internal substructure due to its pairwise bidirectional attention pattern. It is worth noting that although the shortest path distance is leveraged to perform $K$-hop neighbor extraction in this example, several works\citep{abu2019mixhop,chien2020adaptive} leverage graph diffusion kernel to extract $K$-hop neighbors following similar uni-direction hub-spoke attention pattern, and therefore still lacking the ability to learn representations that reflect the internal substructure. 

In this work, we pose the following question: 
\textbf{\textit{How to design appropriate mechanisms to enhance the ability of $K$-hop message-passing GNN to learn node representations that can reflect its internal substructure, and improve its expressive power?}}

Our main contribution is summarized as follows:
\begin{itemize}[topsep=1pt]
    \item We propose a general notion of \textit{substructure encoding function} by establishing connections between $K$-hop message-passing GNNs and 1-WL subgraph GNNs, which may inspire future works to design more powerful and efficient structural encoding schemes.
    \item We propose an instantiation of the substructure encoding function by leveraging multi-step random walk. Our proposed structural encoding scheme is effective, parallelizable with theoretical guarantees to extract structural information of subgraphs. The proposed structural encoding scheme is also orthogonal to previous methods in the literatures.
    \item We propose SEK 1-WL test, which is provably more powerful than Subgraph 1-WL test and $K$-hop 1-WL test, and is not less powerful than 3-WL test.
    \item We propose an implementation of SEK 1-WL test, namely SEK-GNN, which is easy to implement, and can be naturally incorporated into message passing framework with parallelizability and effectiveness. Furthermore, SEK-GNN is able to achieve SOTA performance in a variety of datasets with significantly lower space complexity than 1-WL subgraph GNNs.
\end{itemize}

\section{Preliminary}
\subsection{Notation}
We use $\left\{ \right\}$ to denote sets and $\left\{\left\{\quad \right\}\right\}$ to denote multisets. The index set is denoted as $[n]:=\{1, \cdots, n\}$. Throughout this paper, we consider simple undirected graphs $G=(\mathcal{V}, \mathcal{E})$, where $\mathcal{V}= \left\{v_{1}, \ldots, v_{n}\right\}$ is the node set and $\mathcal{E} \subseteq \mathcal{V} \times \mathcal{V}$ is the edge set. For a node $u$, denote its neighbors as $\mathcal{N}_G(u):=\{v \in \mathcal{V}:\{u, v\} \in \mathcal{E}\}$. The $K$-hop neighbors of node $u$ in graph $G$ are defined as $\mathcal{N}_G^k(u):=\{v \in \mathcal{V}:dis(u,v)=k\}$, where $dis$ is the distance metric to extract k-hop neighbors, e.g., the shortest path distance or random walk steps from node $u$ to node $v$. Let us further define a node set $\mathcal{S}_u^K:=\{u \cup \mathcal{N}_G^k(u): k \in [K]\}$, $G_u^K:=G[\mathcal{S}_u^K]$ is the node-induced subgraph where the central node is $u$, or namely a $K$-hop ego-network rooted at node $u$. Finally, we use $p$ to denote a scalar value, $\textbf{p}$ to denote a column vector, and $P$ to denote a matrix.
\vspace{-3pt}
\subsection{Weisfeiler-Lehman Test}
WL test is a family of very successful algorithmic heuristics used in graph isomorphism problems. 1-WL test, being the simplest one in the family, works as follows--each node is assigned the same color initially, and gets refined in each iteration by
aggregating information from their neighbors' states. The refinement stabilizes after a few iterations and the algorithm outputs a representation of the graph. Two graphs with different representations are not isomorphic. The test can uniquely identify a large set of graphs up to isomorphism~\citep{babai1979canonical}, but there are simple examples where the test tragically fails--for instance, two regular graphs with the same number of nodes and same degrees cannot be distinguished by the test. As a result, a natural extension to 1-WL test is \textit{k}-WL test which provides a hierarchical testing process by keeping the state of \textit{k}-tuples of nodes.
The procedure of 1-dimensional Weisfeiler-Lehman Test algorithm is described in Algorithm 1, where $\mathcal{C}$ denotes the node color space.
\\

\begin{table}
\begin{tabular}{l}
\hline Algorithm 1: The 1-dimensional Weisfeiler-Lehman Algorithm \\
\hline Input : Graph $G=(\mathcal{V}, \mathcal{E})$ and the number of iterations $T$ \\
Output: Color mapping $\chi_G: \mathcal{V} \rightarrow \mathcal{C}$ \\
1 Initialize: Pick a fixed (arbitrary) element $c_0 \in \mathcal{C}$, and \\ let $\chi_G^0(v):=c_0$ for all $v \in \mathcal{V}$ \\
2 for $t \leftarrow 1$ to $T$ do \\
3 $\quad$ for each $v \in \mathcal{V}$ do \\
4 $\quad \quad \chi_G^t(v):=\operatorname{hash}\left(\chi_G^{t-1}(v),\left\{\left\{\chi_G^{t-1}(u): u \in \mathcal{N}_G(v)\right\}\right\}\right)$ \\
5 Return: $\chi_G^T$ \\ \hline
 \end{tabular} 
\end{table}

\subsection{More expressive GNNs}
Despite of the success of MPGNNs, it has been proved in recent literatures that the expressive power of MPGNNs is upper bounded by 1-WL test~\citep{xu2018powerful,morris2019weisfeiler}. Since then numerous approaches for more expressive GNNs have been proposed, including positional and structural encodings~\citep{abboud2020surprising,bouritsas2022improving,puny2020global,dwivedi2021graph,kreuzer2021rethinking,lim2022sign}, higher-order message-passing schemes~\citep{morris2019weisfeiler,maron2020learning,bodnar2021weisfeiler}, equivariant models~\citep{kondor2018covariant,maron2018invariant,maron2019provably,vignac2020building,thiede2021autobahn,de2020natural}

\textbf{Subgraph GNNs}. Recently a collection of methods exploit similar ideas to utilize subgraphs through the application of GNNs. \cite{bevilacqua2021equivariant} explicitly formulated the concept of bags of subgraphs generated by a predefined policy and studied layers to process them in an equivariant manner: the same GNN can encode each subgraph independently (DS-GNN), or information can be shared between these computations in view of the alignment of nodes across the bag \cite{maron2020learning}. Reconstruction GNNs  \cite{cotta2021reconstruction} obtain node-deleted subgraphs, process them with a GNN, and then aggregate the resulting representations by means of a set model. Nested GNN \cite{zhang2021nested} and GNN-As-Kernel \cite{zhao2021stars} propose to hash rooted subgraphs instead of rooted subtrees to perform color refinement. ID-GNN\cite{you2021identity} also leverage ego-network subgraphs with root nodes marked so as to specifically alter the exchange of messages involving them. 

\textbf{$K$-hop message-passing GNNs}. Several works utilize $K$-hop message passing to improve the expressive power of GNNs. \cite{abu2019mixhop} utilizes graph diffusion kernel to extract multi-hop neighbors and calculates the final representation from neighbors from each hop. \cite{nikolentzos2020k} also proposes a $K$-hop GNN model that iteratively updates node representations by aggregating information from $K$-hop neighbors, which is shown to be able to identify fundamental graph properties such as connectivity and bipartiteness. \cite{wang2020multi} proposes to learn attention-based edge coefficients by incorporating information from farther away nodes by means of their shortest path. 

\section{How to design the substructure encoding function}
\label{sec:3}
In the previous section, we show that intuitively $K$-hop message-passing GNNs are less powerful than subgraph 1-WL GNN due to its lacking the ability to capture the internal substructure of the induced subgraph $G_u^K$. We first give a formal definition of \textit{internal substructure}.

\textbf{Definition 1.} (\textbf{\textit{Internal substructure}}) Given a graph $G$ and node $u \in G$, the internal substructure of a $K$-hop node-induced subgraph $G_u^K$ can be represented as a edge set $\mathcal{I}_{G_u^K}$ and the edge induced subgraph $G[\mathcal{I}_{G_u^K}]$, where 

\centerline{$\mathcal{I}_{G_u^K}:=\{\left\{i,j\right\}: (i,j) \in \mathcal{E}, i \in G_u^K, j \in G_u^K, i \neq u, j \neq u \}$.}

Here $\mathcal{I}_{G_u^K}$ refers to set of edges in $G_u^K$ that excludes node $u$. As $K$-hop message-passing GNNs are not able to be aware of the internal substructure, we propose to use a \textit{substructure encoding function} to encode the information of node $u$'s internal substructure of $G_u^K$. Next, we give a formal definition of \textit{substructure encoding function}.

\textbf{Definition 2.} (\textbf{\textit{Substructure encoding function}}) A substructure encoding function $f: G_u^K \times G \rightarrow \mathcal{R}^d$ takes as input a $K$-hop subgraph rooted at node $u$ and graph $G$, and outputs an encoded $d$-dimensional features which can reflect the internal substructure of $G_u^K$. 

By designing a proper substructure encoding function $f(\cdot)$ and incorporating the encoded information, the expressive power of $K$-hop message-passing GNNs will get uplifted. One core question in this work is \textit{how to design a proper substructure encoding function $f$ to enhance the expressive power of a $K$-hop message-passing GNNs}. $f(\cdot)$ is expected to be able to encode the internal substructure with computational efficiency in terms of space and time complexity. One way to design such a function is to exploit random walk to calculate the self-return probability for every node $u \in \mathcal{V}$. Intuitively, two nodes with different internal substructures would lead to different random walk patterns given sufficient steps of random walk and the self-return probability $(p_u^1,p_u^2,\cdots p_u^l) \neq (p_v^1,p_v^2,\cdots p_v^l)$ for node $u$ in graph $G$ and node $v$ in graph $H$, if the internal substructure of $G_u^K$ is not identical to $H_v^K$. Here $(p_u^1,p_u^2,\cdots p_u^l)$ is the $l$-step self-return probability for node $u$. But how many steps are sufficient for node $u$ and $v$ to distinguish the internal substructure of $G_u^K$ and $H_v^K$? We now propose the following theorem to give a lower bound for the steps needed to encode the internal substructure of a $K$-hop node-induced subgraph $G_u^K$. 

\begin{theorem}
\label{theorem-1}
Given two $n$-node $r$-regular graphs $G$ and $H$, let $3 \leq r<(2 \log 2 n)^{1 / 2}$ and $\epsilon$ be a fixed constant. For two $K$-hop ego-networks $G_u^K$ and $H_v^K$ with $K$ being at most $\left\lceil\left(\frac{1}{2}+\epsilon\right) \frac{\log 2 n}{\log (r-1)} + 1 \right\rceil$, $2K$ steps of random walk is sufficient to discriminate the internal substructure of $G_u^K$ and $H_v^K$.
\end{theorem}

The proof is included in Appendix \ref{proof theo1}. Theorem \ref{theorem-1} demonstrates the advantage of using the self-return probability of random walk as the substructure encoding function $f(\cdot)$, thanks to its computational efficiency as it only requires $2K$ steps to encode the internal substructure of a $K$-hop ego-network. Although theorem \ref{theorem-1} only applies to regular graphs, empirically we demonstrate that $\left\{p_u^i\right\}_{i=1}^l$ is able to lead to good model performance with a moderately small integer $l$. Furthermore, the computation of the self-return probability is easy to achieve parallelism using matrix computations. Let $\widetilde{A}= A+I$ be the adjacency matrix with self edges, and $\widetilde{D}$ be the corresponding diagonal matrix. $\mathbf{p}_u^t=\widetilde{D}^{-1}\widetilde{A}\mathbf{p}_u^{t-1}$, where $\mathbf{p}_u^t$ is a column vector to represent node $u$'s landing probability to every node $v \in \mathcal{V}$ at time step $t$, denoted by $\mathbf{p}_u^t[v]$. This equation can be expressed in matrix form as $H^{(t+1)}=\widetilde{D}^{-1}\widetilde{A}H^{(t)}$, where $H^{\left(0\right)}=I$.  $\left\{H^{(t)}\right\}_{t=1}^l$ now contains the self-return probability for every node $u \in \mathcal{V}$, which can be obtained as $\left\{H_{uu}^{(t)}\right\}_{t=1}^l$ for $l$ steps of random walk starting from node $u$.

\textbf{Injecting more substructure information.} Although self-return probability $\left\{H_{uu}^{(t)}\right\}_{t=1}^l$ is able to encode the internal substructure of a $K$-hop ego-network $G_u^K$ for node $u$ in graph $G$, $\left\{H^{(t)}\right\}_{t=1}^l$ actually contains more structural information that can be utilized. Therefore, we propose another two functions to encode more substructure information. \textit{(i). Central node to $k$-hop neighbors landing probability} is to calculate an aggregated statistics of the landing probability from central node $u$ to $k$-hop neighboring nodes $\mathcal{N}_G^k(u):k \in [K]$ for each time step $t$. \textit{(ii). Landing probability across $k$-hop neighbors} is to calculate an aggregated statistic of the nodes that are equally distant from root node $u$. Let $f_1(\cdot)$ denote the function to encode the self-return probability, $f_2(\cdot)$ denotes the function to encode the aggregated landing probability of a central node to $K$-hop neighbors, and $f_3(\cdot)$ is the function to encode the aggregated landing probability across $k$-hop neighbors for all $K$. Formally, we have the following substructure encoding function:
\begin{equation}
\begin{aligned}
&f_1(G_u^K,G) := \left\{H_{uu}^{(t)}\right\}_{t=1}^l, \\
&f_2(G_u^K,G) := \operatorname{AGG}_k(\left\{H^{(t)}_{ui}:dis(u,i)=k,k \in [K] \right\}_{t=1}^l), \\
&f_3(G_u^K,G) := \operatorname{AGG}_k(\left\{H^{(t)}_{i,j}:dis(i,u)=dis(j,u),k \in [K] \right\}_{t=1}^l), \\
&f(G_u^K,G) := \operatorname{COMBINE}({f_i(G_u^K,G),i \in \left\{1,2,3\right\}}). 
\end{aligned}
\label{Eq:encoding}
\end{equation}

For $l$-step random walk, the encoding function $f(\cdot)$ will output a $l$-dimensional features to encode the internal substructure of node $u \in \mathcal{V}$, extracted from $\left\{H^{(t)}\right\}_{t=1}^l$. 

\textbf{Injecting contextualized substructure information.} Thanks to the substructure encoding function $f(\cdot)$, we can encode the internal substructure of $G_u^K$ for every node $u \in \mathcal{V}$. Node representation with its enclosing $K$-hop ego-network $h_v := h_{G_v^K}$ can be further enriched by incorporating the substructure information of node $v$, which is $\left\{\left\{h_u: u \in G_v^K, u \neq v\right\}\right\}$. This can be easily achieved via $f(G_u^K,G)$ as node features. 

\section{Substructure Enhanced $K$-hop 1-WL Algorithm}
In this section, we first generalize previous works on $K$-hop message-passing GNNs by presenting a formal definition of the $K$-hop 1-WL color refinement algorithm, as well as subgraph 1-WL color refinement algorithm which was proposed in previous works\cite{zhang2021nested,zhao2021stars}. Then we propose our novel color refinement algorithm, namely the Substructure Enhanced $K$-hop 1-WL Algorithm (SEK 1-WL), which is more expressive than $K$-hop 1-WL and subgraph 1-WL. Finally, we give a practical implementation of SEK 1-WL which is efficient and easy to implement.

\textbf{Definition 3.} (\textbf{\textit{$K$-hop 1-WL Test}}) A $K$-hop 1-WL color refinement algorithm iteratively refines node colors using all $k$-hops neighbors, where $k \in [K]$. Formally, the color refinement algorithm is 
\begin{equation}
\label{eq:k_hop}
\begin{aligned}
\chi_G^t(v) := \operatorname{hash}\Bigg(\chi_G^{t-1}(v), & \Big\{\Big\{\chi_G^{t-1}(u) : u \in \mathcal{N}_G(v)\Big\}\Big\}, \\
& \Big\{\Big\{\chi_G^{t-1}(u) : \operatorname{dis}_G(v, u) = 2\Big\}\Big\}, \\
& \cdots, \\
& \Big\{\Big\{\chi_G^{t-1}(u) : \operatorname{dis}_G(v, u) = K\Big\}\Big\}\Bigg).
\end{aligned}
\end{equation}

The only difference between 1-WL test and $K$-hop 1-WL test lies in line 4 of Algorithm 1, where the color refinement update procedure is replaced by Eq. \ref{eq:k_hop}. The  $K$-hop 1-WL algorithm generalizes previous works on $K$-hop message-passing GNNs where $dis$ can be the shortest path distance using shortest path kernel\cite{brossard2020graph,nikolentzos2020k} or random walk steps using graph diffusion kernel\cite{abu2019mixhop,wang2020multi}. Next, we introduce the definition of the Subgraph 1-WL test which has been realized in previous works\cite{zhao2021stars,zhang2021nested}.

\textbf{Definition 4.} (\textbf{\textit{Subgraph 1-WL Test}}) A subgraph 1-WL color refinement algorithm hashes the node-induced subgraph instead of direct neighbors. Formally, the color refinement algorithm is as follows: 
\begin{equation}
\label{eq:subgraph_wl}
\begin{aligned}
\chi_G^t(v):=\operatorname{hash}(\chi_G^{t-1}(v),\gamma_G^{t-1}(G_v^K)),
\end{aligned}
\end{equation}

where $\gamma_G: G_v^K \rightarrow \mathcal{C}$ is a function that maps a (sub)graph to a color $c \in \mathcal{C}$. As the color map for graphs is as hard as graph isomorphism, previous works realize Subgraph 1-WL test by replacing the hashing function $\gamma$ with a GNN model as a wrapper followed by a graph readout function to obtain the subgraph representation of $G_v^K$, and then
the whole graph representation is then obtained by pooling these subgraph representations \cite{zhang2021nested,zhao2021stars}. 

\textbf{Definition 5.} (\textbf{\textit{SEK 1-WL Test}}) Finally our proposed color refinement algorithm SEK 1-WL updates node colors using both $K$-hop neighbors as well as the contextualized internal substructure information, and the color refinement procedure to replace line 4 of Algorithm 1 is shown as follows:
\begin{equation}
\label{eq:sek}
\begin{aligned}
\chi_G^t(v) := \operatorname{hash}\Bigg(
    & \chi_G^{t-1}(v), f(G_v^K, G), \Big\{\Big\{f(G_u^K, G) : u \in G_v^K, u \neq v\Big\}\Big\}, \\
    & \Big\{\Big\{\chi_G^{t-1}(u) : u \in \mathcal{N}_G(v)\Big\}\Big\}, \\
    & \Big\{\Big\{\chi_G^{t-1}(u) : \operatorname{dis}_G(v, u) = 2\Big\}\Big\}, \\
    & \cdots, \\
    & \Big\{\Big\{\chi_G^{t-1}(u) : \operatorname{dis}_G(v, u) = K\Big\}\Big\}
\Bigg).
\end{aligned}
\end{equation}

In Eq. \ref{eq:sek}, $f(\cdot)$ is the substructure encoding function proposed in Eq. \ref{Eq:encoding}. As shown in Eq. \ref{eq:sek}, SEK 1-WL utilizes $K$-hop neighborhood information, similarly to $K$-hop 1-WL test and also $f(G_v^K,G)$, the encoded features of internal substructure of node $v$. 
Furthermore, SEK 1-WL also inject contextualized substructure information  $\left\{\left\{f(G_u^K,G): u \in G_v^K, u \neq v \right\}\right\}$ for node $v$ to enhance the expressive power. We now propose a theorem regarding the expressive power of SEK 1-WL.

\begin{theorem}
\label{theo-2}
SEK 1-WL test is strictly more powerful than $K$-hop 1-WL test and Subgraph 1-WL test.
\end{theorem}

\textit{Proof.} It is easy to see that SEK 1-WL is more powerful than $K$-hop 1-WL test as SEK 1-WL incorporates additional substructure information using $f(\cdot)$, and both of them perform node color refinement using $K$-hop neighbors. Secondly, we show that SEK 1-WL test is more powerful than Subgraph 1-WL test based on the observation of the attention patterns induced by $K$-hop message-passing GNN and 1-WL subgraph GNN discussed in section \ref{intro}: hashing a subgraph $G_v^K$ can be equivalently expressed as hashing its $K$-hop neighbors as well as its internal substructure of root node $v$, and then the color refinement algorithm of subgraph 1-WL test can be reformulated as follows:
\begin{equation}
\label{eq:subgraph 1-wl v2}
\begin{aligned}
\chi_G^t(v) := \operatorname{hash}\Bigg(
    & \chi_G^{t-1}(v), f(G_v^K, G), 
    \Big\{\Big\{\chi_G^{t-1}(u) : u \in \mathcal{N}_G(v)\Big\}\Big\}, \\
    & \Big\{\Big\{\chi_G^{t-1}(u) : \operatorname{dis}_G(v, u) = 2\Big\}\Big\}, \\
    & \cdots, \\
    & \Big\{\Big\{\chi_G^{t-1}(u) : \operatorname{dis}_G(v, u) = K\Big\}\Big\}
\Bigg).
\end{aligned}
\end{equation}

The extra expressive power of SEK 1-WL over subgraph 1-WL stems from the contextualized substructure information 

$\left\{\left\{f(G_u^K,G): u \in G_v^K, u \neq v \right\}\right\}$
injected in the color refinement procedure of SEK 1-WL. Finally, it is easy to see that SEK 1-WL is more powerful than 1-WL test. This concludes the proof of Theorem \ref{theo-2}. Next we give a practical implementation of SEK 1-WL.

\textbf{Practical Implementation.} The SEK 1-WL color refinement algorithm can be easily implemented using the message-passing GNN framework. We call our framework SEK-GNN, with the following message-passing and update step for one iteration:
\begin{equation}
\begin{small}
\begin{aligned}
&{m_v^{l, k}=\operatorname{MESSAGE}_k^l\left(h_v^{l-1},f\left(G_v^K, G\right),\left\{\left\{\left(h_u^{l-1},f\left(G_u^K, G\right)\right): u \in \mathcal{N}_G^k(v)\right\}\right\}\right)}, \\
& h_v^{l, k}=\operatorname{UPDATE}_k^l\left(m_v^{l, k}\right), \\
& h_v^l=\operatorname{COMBINE}^l\left(\left\{\left\{h_v^{l, k}: k \in[K]\right\}\right\}\right).  \\ 
\end{aligned}
\end{small}
\label{eq:sek-gnn}
\end{equation}

SEK-GNN follows a similar message passing and update procedure as the previous $K$-hop message-passing GNN, where $h_v^{l, k}$ is the representation of node $v$ at $l^{th}$ layer and $k^{th}$ hop. Once $h_v^{l, k}$ is calculated using the $MESSAGE$ and $UPDATE$ functions, a $COMBINE$ function is leveraged to combine the node $v$'s representation from $k$ hop, for all $k \in [K]$. However, the difference between our work and the previous ones is that for each node $v \in \mathcal{V}$, both $f\left(G_u^K, G\right)$ and the contextualized encoded substructure features $\left\{\left\{f(G_u^K,G): u \in G_v^K, u \neq v \right\}\right\}$ are involved in the $MESSAGE$ function. Some design choices for the $MESSAGE$ function include GCN\cite{kipf2016semi} and GIN\cite{xu2018powerful}; for a different $k$-hop message passing, $MESSAGE$ function can share the same model parameters or use different parameters for each $k$. $UPDATE$ function can be a MLP or non-linear activation function such as RELU. We also leverage jumping knowledge network\cite{xu2018representation} to increase the model capacity, and the pooling function can be summation, concatenation, or attention-based aggregation. Finally, for $COMBINE$ operation, summation or aggregation following geometric distribution are considered. For $geometric$, the weight of hop $k$ is calculated based on $\theta_k=\alpha(1-\alpha)^k$, where $\alpha \in(0,1]$. The final representation for all $k$ hop nodes is obtained via weighted sum. SEK-GNN, as an instance of SEK 1-WL, is more expressive than $K$-hop 1-WL test and Subgraph 1-WL test. 

\begin{figure*}[h]
\centering
\includegraphics[width=0.7\textwidth]{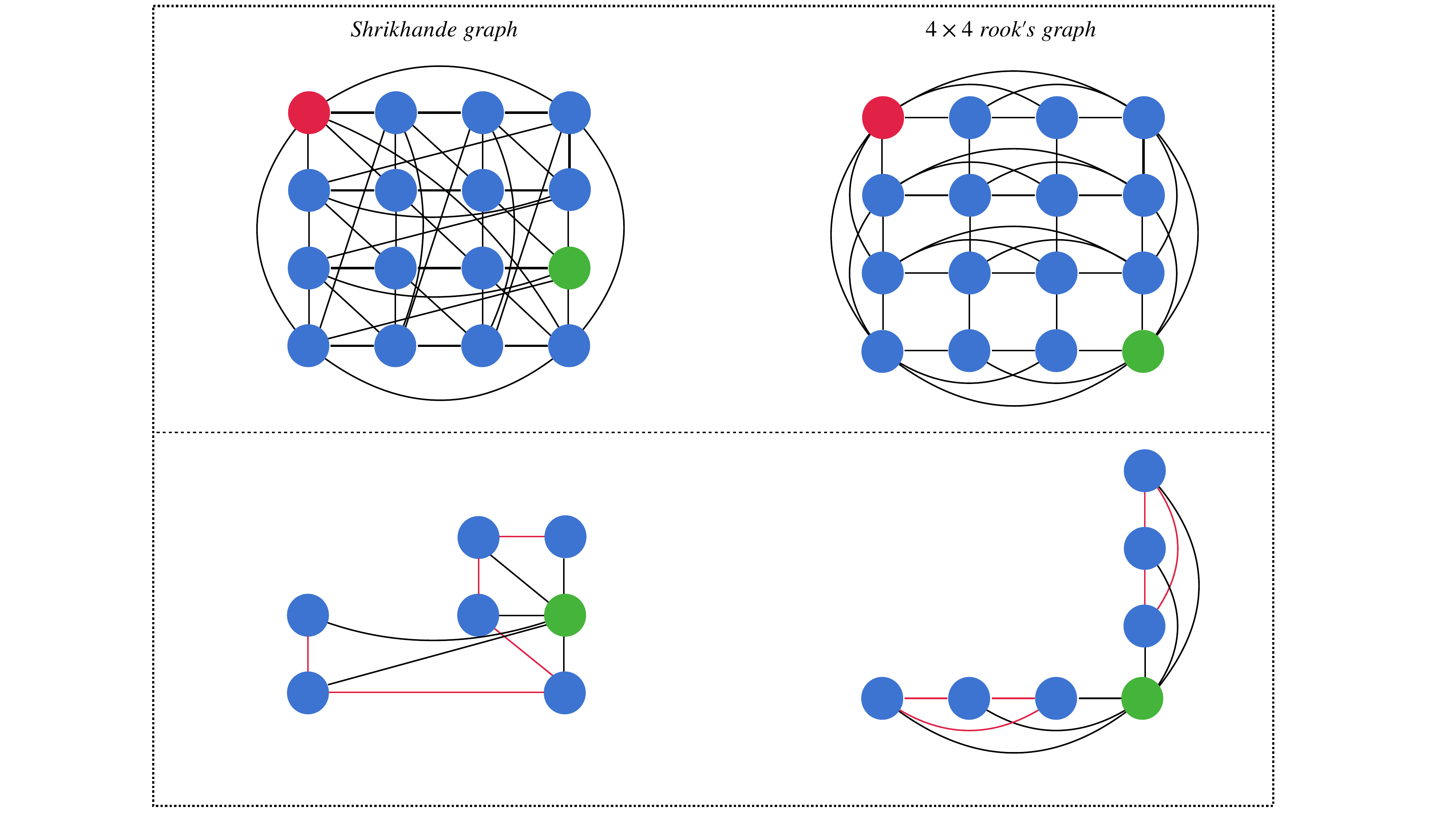}
\caption{Two non-isomorphic graphs with the same intersection array, $\left\{6,3;1,2\right\}$; for the two \textcolor{red}{red} nodes, we show one of the 2-hop neighbors highlighted in \textcolor{green}{green}, and their corresponding 1-hop induced subgraphs.}
\label{fig:distance-regular-graph}
\end{figure*}

\begin{prop}
SEK-GNN is not less powerful than 3-WL test.
\end{prop}

\cite{frasca2022understanding} has shown that all subgraph-based GNNs are bounded by 3-WL test by proving that all such methods can be implemented by 3-IGN\citep{maron2019provably,maron2018invariant}, and \cite{feng2022powerful} shows that $K$-hop message-passing GNNs can be implemented by 3-IGN; therefore SEK-GNN, as a $K$-hop message-passing GNN framework, is also upper bounded by 3-WL test. We say that A is not less powerful than B if there exists a pair of non-isomorphic graphs that cannot be distinguished by B but can be distinguished by A\cite{chen2020can}. Figure \ref{fig:distance-regular-graph} illustrates two well-known non-isomorphic distance-regular graphs: the $4 \times 4$ rook's graph and the Shrikhande graph, both of which have the same intersection array $\left\{6,3;1,2\right\}$ (the definition of distance-regular graph and intersection array is introduced in Appendix \ref{definition}). As both graphs are strongly regular, 3-WL test fails to discriminate them; however, SEK-GNN is able to discriminate them as for the two red nodes, we can see that the 1-hop induced subgraphs of the green nodes (both are 2-hop neighbors of red nodes) have different internal substructures: for the $4 \times 4$ rook's graph, there are two circles in the subgraph induced by blue nodes; however, for the Shrikhande graph, there is no circle involved in the subgraph induced by blue nodes. This demonstrates the effectiveness of incorporating contextualized substructures. One could also verify that the 1-hop induced subgraphs for the two red nodes also have different internal substructures. 

\section{Space and time complexity}

As SEK-GNN involves a preprocessing stage and a training (inference) stage, we discuss space and time complexity for both stages separately.

For a graph $G=(\mathcal{V},\mathcal{E})$ with $n$ nodes and $m$ edges, in the preprocessing stage, to compute $\left\{H^{(t)}\right\}_{t=1}^l$ for a $l$-step random walk, SEK-GNN only requires $\mathcal{O}(2m)$ space since the calculation of $H^{(t)}$ only depends on $H^{(t-1)}$, which is on par with standard GNN models such as GCN and GIN. In the training and inference stage, SEK-GNN only requires $\mathcal{O}(n)$ space as the substructure encoding function outputs $f(G_v^K,G)$ for every node $v$ as the node features. The space complexity of SEK-GNN is similar to that of previous works on $K$-hop GNNs, and is significantly less than 1-WL subgraph GNN models, which requires $\mathcal{O}(n^2)$ due to the independent subgraph extraction.

For time complexity, SEK-GNN requires $\mathcal{O}(lm)$ time to preprocess $\left\{H^{(t)}\right\}_{t=1}^l$ for a $l$-step random walk. As we show in section 3, $l$ is typically a small integer independent of graph size and the preprocessing of $\left\{H^{(t)}\right\}_{t=1}^l$ only performs once, hence it is still affordable. For the training and inference stage, SEK-GNN requires $\mathcal{O}(n^2)$ time in the worst case. However, empirically, we only randomly sample a fixed number of neighbors from $\mathcal{N}_G^K(v)$ for every node $v$, and do not see a drop in model performance. With neighbor sampling, SEK-GNN only requires $\mathcal{O}(cn)$ time, where $c$ is a constant only dependent on the number of neighbors to sample for each hop $k \in [K]$. The time complexity of SEK-GNN is also clearly less than 1-WL subgraph GNN models whose time complexity is $\mathcal{O}(nm)$, note that typically $c \ll m$. 

\section{Related Work}

Several works also utilize random walk to enhance the expressivity of graph neural networks \citep{xu2018representation,page1999pagerank,nikolentzos2020random,klicpera2018predict,Liu_2020}. \cite{xu2018representation} made the observation that the range of "neighboring" nodes that a node’s representation draws from strongly depends on the graph structure, analogous to the spread of a random walk. Motivated by this observation, they proposed a new aggregation scheme for node representational learning that can adapt neighborhood ranges to nodes individually. \cite{klicpera2018predict} established connections between graph convolutional networks (GCN) and PageRank \cite{page1999pagerank} to derive an improved propagation scheme based on personalized PageRank. Random walk GNN \cite{nikolentzos2020random} uses a number of trainable "hidden graphs" which are compared against the input graphs using a random walk kernel to produce graph representations. Although these works leverage random walk to build new GNN architectures, SEK-GNN is inherently different from these works, which leverages the sequential information $\left\{H^{(t)}\right\}_{t=1}^l$ of a $l$-step random walk to encode the internal substructure of a induced subgraph to enhance the representative power of $K$-hop message-passing GNNs.

Another line of research tries to enhance the expressive power of GNNs by augmenting node features. \cite{sato2021random,loukas2019graph,abboud2020surprising,dasoulas2019coloring} add one-hot or random features into node labels, although being scalable and easy to implement, these approaches may not preserve permutation invariant property, thus would hurt the generalization performance
outside of the training set. Recent work also proposes to use distance features to uplift GNN’s expressivity~\cite{li2020distance}
where a distance vector w.r.t the target node set is calculated for each node as its additional feature. \cite{zhang2021labeling} proposes labelling trick to perform node feature augmentation by distinguishing target node set. \cite{wijesinghe2021new} introduces a hierarchy of local isomorphism and proposes structural coefficients as additional features to identify such local isomorphism. Our proposed substructure encoding function also belongs to this line of works, where \textit{self-return probability}, \textit{landing probability across $k$-hop neighbors} and \textit{central node to $k$-hop neighbors landing probability} are calculated to enrich node features, which can extract structural disparity with theoretical guarantees. Our approach is orthogonal to previous structure encoding methods, and provides an easy way to incorporate contextualized substructure information. 

Finally, \cite{feng2022powerful} also aims to improve the expressive power of $K$-hop message-passing GNNs. However, they leverage peripheral edges to enrich the representation of a $K$-hop message-passing GNNs, which is equivalent of extracting the internal substructure of $G_u^K$ explicitly while SEK-GNN leverages substructure encoding function to encode the internal substructure implicitly. However, as SEK-GNN also injects contextualized substructure information $\left\{\left\{f(G_v^K,G): v \in G_u^K, v \neq u \right\}\right\}$, the expressive power can get further improved. 

\begin{figure*}
\centering
\includegraphics[width=0.7\textwidth]{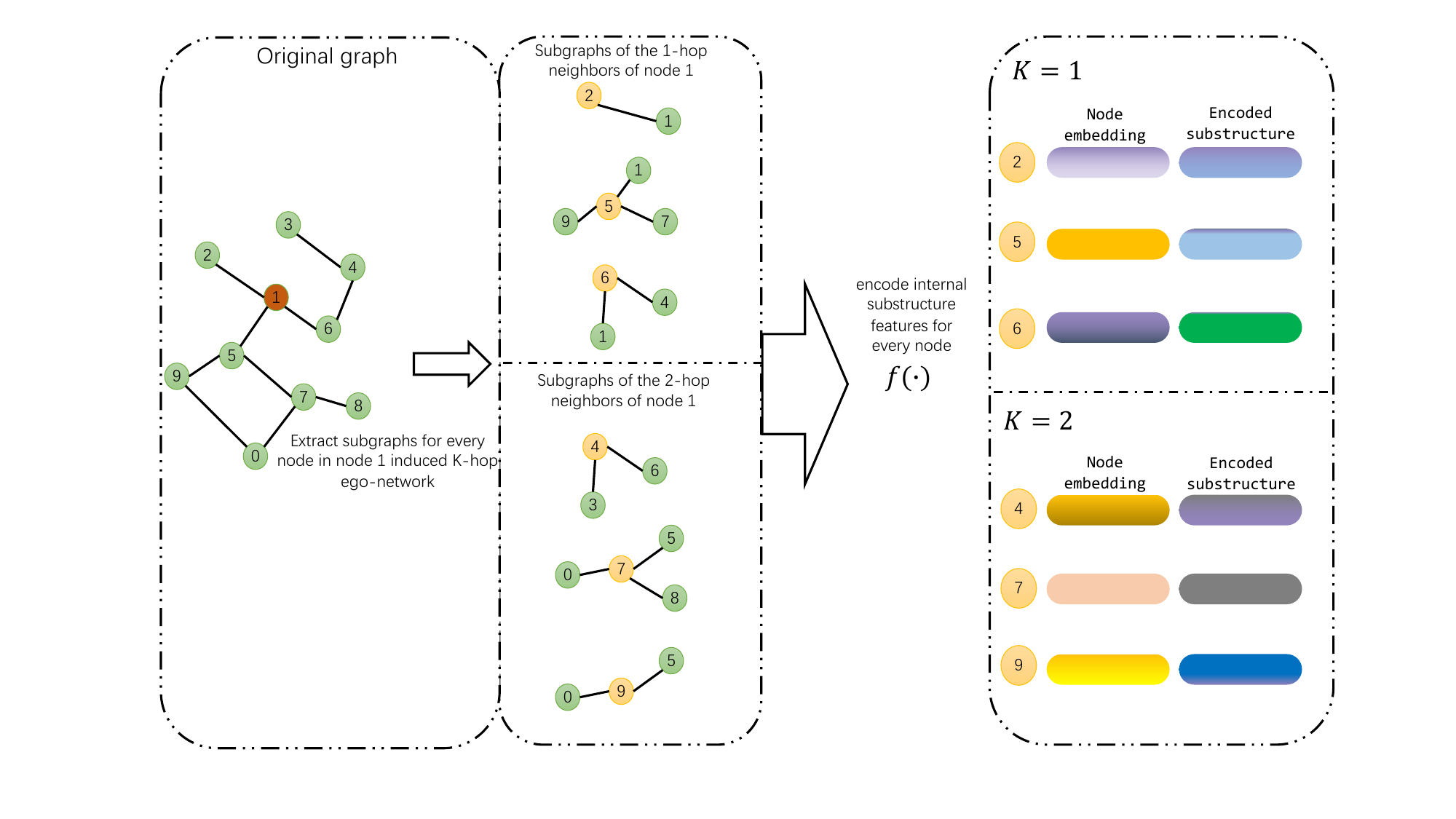}
\caption{A toy example to illustrate how to inject contextualized substructure information to enrich the representation in SEK-GNN. Specifically, in this example the $K$-hop neighbors of node 1 are extracted for all $k \in [K]$ ($K=2$ in this case, and ranges between 3 and 6 in the experiments), and then for each node, we further extract its $h$-hop induced subgraphs ($h=1$ in this example and is typically 3 to 6 in the experiments) and calculate the substructure information using $f(\cdot)$. Then for each $k$-hop neighbors of node 1 where $k \in [K]$, the node representation consists of two parts: i) node embedding from the previous iteration and ii) encoded substructure features using $f(\cdot)$. Finally, node 1's representation is updated according to Equation \ref{eq:sek-gnn}. Please also note that the induced subgraph of node 1 is also extracted and encoded using $f(\cdot)$ which is omitted in this figure.}
\label{fig:sek-gnn}
\end{figure*}

\section{Experiments}
In this section, we empirically evaluate the proposed method on synthetic datasets and real-world datasets with both graph classification and graph regression tasks. We demonstrate that SEK-GNN can achieve state-of-the-art performance over multiple datasets. We make use of Pytorch Geometric~\citep{fey2019fast} to implement our proposed framework, and our code is available at \url{https://github.com/tianyao-aka/Expresive_K_hop_GNNs}. The statistics of the datasets used in the experiment are shown in Appendix \ref{appendix:dataset_stats}.

\subsection{Synthetic datasets}
\textbf{Datasets.} We use two synthetic datasets to evaluate the effectiveness of SEK-GNN: i) Graph property regression, where the goal is to regress the number of three graph properties on random graphs, namely connectedness, diameter, radius\cite{corso2020principal}; ii) Graph substructure counting, where the goal is to regress the number of four substructures: triangle, tailed triangle, star, and 4-cycle on randomly generated graphs\cite{chen2020can}.

\textbf{Baseline methods.} We use GIN \cite{xu2018powerful} as a baseline method as it has the same expressive power as 1-WL test. For more powerful baselines, we use GIN-AK+ \cite{zhao2021stars}, PNA \cite{corso2020principal}, PPGN \cite{maron2019provably}, and KP-GIN+  \cite{feng2022powerful}.

\begin{table*}[htp]
\begin{tabular}{@{}l|lll|llll@{}}
\toprule
\multicolumn{1}{c|}{} & \multicolumn{3}{c|}{Graph Properties}                  & \multicolumn{4}{c}{Counting Substructures}                                  \\ \cmidrule(l){2-8} 
\multicolumn{1}{c|}{\multirow{-2}{*}{Method}} &
  {\color[HTML]{413D44} Connect.} &
  {\color[HTML]{413D44} Diameter} &
  {\color[HTML]{413D44} Radius} &
  {\color[HTML]{413D44} Tri.} &
  {\color[HTML]{413D44} Tailed Tri.} &
  {\color[HTML]{413D44} Star} &
  {\color[HTML]{413D44} 4-Cycle} \\ \midrule
GIN                   & -1.9239          & -3.3079          & -4.7584          & 0.3569          & 0.2373                                 & 0.0224          & 0.2185 \\ \midrule
PNA                   & -1.9395          & -3.4382          & -4.947           & 0.3532          & 0.2648                                 & 0.1278          & 0.243  \\
PPGN                  & -1.9804          & -3.6147          & -5.0878          & 0.0089          & 0.0096                                 & 0.0148          & 0.009  \\
GIN-AK+               & -2.7513          & -3.9687          & -5.1846          & 0.0123          & 0.0112                                 & 0.015           & 0.0126 \\
K-GIN+                & -2.1782          & \textbf{-3.9695} & -5.3088          & 0.1316          & 0.0813                                 & 0.0017          & 0.0916 \\
KP-GIN+               & \textbf{-4.4322} & -3.9361          & \textbf{-5.3345} & \textbf{0.0012} & {\color[HTML]{FE0000} \textbf{0.0016}} & \textbf{0.0014} & 0.0067 \\ \midrule
SEK-GIN &
  -2.6162 &
  -3.6101 &
  {\color[HTML]{FE0000} \textbf{-5.6821}} &
  0.0031 &
  0.0058 &
  {\color[HTML]{FE0000} \textbf{0.0007}} &
  {\color[HTML]{FE0000} \textbf{0.0032}} \\
SEK-PPGN &
  {\color[HTML]{FE0000} \textbf{-4.9883}} &
  {\color[HTML]{FE0000} \textbf{-4.4438}} &
  -5.2273 &
  {\color[HTML]{FE0000} \textbf{0.0010}} &
  \textbf{0.0021} &
  0.0020 &
  \textbf{0.0063} \\ \bottomrule
\end{tabular}
\caption{Simulation dataset results. Top two results are highlighted by \textcolor{red}{\textbf{First}} and \textbf{Second}.}
\label{exp:syn}
\end{table*}

\begin{table*}[h]
\scalebox{0.85}{
\begin{tabular}{@{}ccccccccc@{}}
\toprule
{\color[HTML]{363438} \textbf{Method}} &
  {\color[HTML]{363438} \textbf{MUTAG}} &
  {\color[HTML]{363438} \textbf{PTC-MR}} &
  {\color[HTML]{363438} \textbf{PROTEINS}} &
  \textbf{ENZYMES} &
  {\color[HTML]{363438} \textbf{BZR}} &
  {\color[HTML]{4D4852} \textbf{COX2}} &
  {\color[HTML]{363438} \textbf{IMDB-B}} &
  {\color[HTML]{363438} \textbf{IMDB-M}} \\ \midrule
{\color[HTML]{363438} WL} &
  {\color[HTML]{4D4852} 90.4 ± 5.7} &
  {\color[HTML]{4D4852} 59.9 ± 4.3} &
  {\color[HTML]{4D4852} 75.0 ± 3.1} &
  - &
  {\color[HTML]{4D4852} 78.5 ± 0.6} &
  {\color[HTML]{4D4852} 81.7 ± 0.7} &
  {\color[HTML]{4D4852} 73.8 ± 3.9} &
  50.9± 3.8 \\
{\color[HTML]{363438} RetGK} &
  {\color[HTML]{4D4852} 90.3 ± 1.1} &
  {\color[HTML]{4D4852} 62.5 ± 1.6} &
  {\color[HTML]{4D4852} 75.8 ± 0.6} &
  - &
   &
  - &
  {\color[HTML]{4D4852} 71.9 ± 1.0} &
  47.7± 0.3 \\
{\color[HTML]{363438} GNTK} &
  {\color[HTML]{4D4852} 90.0 ± 8.5} &
  {\color[HTML]{363438} \textbf{67.9 ± 6.9}} &
  {\color[HTML]{4D4852} 75.6 ± 4.2} &
  - &
  {\color[HTML]{4D4852} 83.6 ± 2.9} &
  - &
  {\color[HTML]{4D4852} 76.9 ± 3.6} &
  - \\
WWL &
  87.2 ± 1.5 &
  66.3 ± 1.2 &
  {\color[HTML]{4D4852} 74.2 ± 0.5} &
  - &
  {\color[HTML]{4D4852} 84.4 ± 2.0} &
  {\color[HTML]{4D4852} 78.2 ± 0.4} &
  {\color[HTML]{4D4852} 74.3 ± 0.8} &
  - \\
{\color[HTML]{363438} FGW} &
  {\color[HTML]{4D4852} 88.4 ± 5.6} &
  {\color[HTML]{4D4852} 65.3 ± 7.9} &
  {\color[HTML]{4D4852} 74.5 ± 2.7} &
  - &
  {\color[HTML]{4D4852} 85.1 ± 4.1} &
  {\color[HTML]{4D4852} 77.2 ± 4.8} &
  {\color[HTML]{4D4852} 63.8 ± 3.4} &
  - \\ \midrule
{\color[HTML]{363438} DGCNN} &
  {\color[HTML]{4D4852} 85.8 ± 1.7} &
  {\color[HTML]{4D4852} 58.6 ± 2.5} &
  {\color[HTML]{4D4852} 75.5 ± 0.9} &
  38.9±5.7 &
  - &
  - &
  {\color[HTML]{4D4852} 70.0 ± 0.9} &
  47.8 \\
{\color[HTML]{363438} CapsGNN} &
  {\color[HTML]{4D4852} 86.6 ± 6.8} &
  {\color[HTML]{4D4852} 66.0 ± 1.8} &
  {\color[HTML]{4D4852} 76.2 ± 3.6} &
  - &
  - &
  - &
  {\color[HTML]{4D4852} 73.1 ± 4.8} &
  - \\
{\color[HTML]{4D4852} GraphSAGE} &
  {\color[HTML]{4D4852} 85.1 ± 7.6} &
  {\color[HTML]{4D4852} 63.9 ± 7.7} &
  {\color[HTML]{4D4852} 75.9 ± 3.2} &
  - &
  - &
  - &
  {\color[HTML]{4D4852} 72.3 ± 5.3} &
  - \\
{\color[HTML]{4D4852} GIN} &
  {\color[HTML]{4D4852} 89.4 ± 5.6} &
  {\color[HTML]{4D4852} 64.6 ± 7.0} &
  {\color[HTML]{4D4852} 75.9 ± 2.8} &
  59.6±4.5 &
  - &
  - &
  {\color[HTML]{4D4852} 75.1 ± 5.1} &
  \textbf{52.3± 2.8} \\
GIN-AK+ &
  91.3± 7.0 &
  {\color[HTML]{FF0000} \textbf{68.20± 5.6}} &
  {\color[HTML]{363438} \textbf{77.1 ± 5.7}} &
  - &
  - &
  - &
  {\color[HTML]{4D4852} 75.6 ± 3.7} &
  - \\
{\color[HTML]{363438} GraphSNN} &
  {\color[HTML]{363438} 91.57 ± 2.8} &
  {\color[HTML]{4D4852} 66.70 ± 3.7} &
  {\color[HTML]{363438} 76.83 ± 2.5} &
  \textbf{61.7±3.4} &
  {\color[HTML]{363438} \textbf{88.69 ± 3.2}} &
  \textbf{82.86 ± 3.1} &
  {\color[HTML]{FF0000} \textbf{77.86 ± 3.6}} &
  - \\
{\color[HTML]{363438} KP-GIN} &
  \textbf{92.2± 6.5} &
  {\color[HTML]{4D4852} 66.80 ± 6.8} &
  {\color[HTML]{4D4852} 75.80 ± 4.6} &
  - &
  - &
  - &
  {\color[HTML]{4D4852} 76.6 ± 4.2} &
  - \\
{\color[HTML]{363438} SEK-GIN} &
  {\color[HTML]{FF0000} \textbf{92.2± 6.2}} &
  66.86± 7.2 &
  {\color[HTML]{FF0000} \textbf{78.32 ± 2.7}} &
  {\color[HTML]{FF0000} \textbf{61.85±5.1}} &
  {\color[HTML]{FF0000} \textbf{89.65 ± 3.8}} &
  {\color[HTML]{FF0000} \textbf{85.89±4.4}} &
  \textbf{76.9 ± 3.8} &
  {\color[HTML]{FF0000} \textbf{53.80 ± 3.2}} \\ \midrule
†GIN-AK+ &
  95.1± 6.1 &
  74.1± 5.9 &
  {\color[HTML]{363438} 78.9 ± 5.4} &
  - &
  - &
  - &
  {\color[HTML]{4D4852} 77.3 ± 3.1} &
  - \\
{\color[HTML]{363438} †GraphSNN} &
  {\color[HTML]{363438} 94.70 ± 1.9} &
  {\color[HTML]{363438} 70.58 ± 3.1} &
  {\color[HTML]{363438} 78.42 ± 2.7} &
  - &
  {\color[HTML]{363438} \textbf{91.12 ± 3.0}} &
  {\color[HTML]{363438} \textbf{86.28 ± 3.3}} &
  {\color[HTML]{363438} 78.51 ± 2.8} &
  - \\
†KP-GIN &
  \textbf{96.1± 4.6} &
  {\color[HTML]{FE0000} \textbf{76.20 ± 4.5}} &
  {\color[HTML]{4D4852} \textbf{79.50 ± 4.4}} &
  - &
  - &
  - &
  {\color[HTML]{FF0000} \textbf{80.7 ± 2.6}} &
  - \\
†SEK-GIN &
  {\color[HTML]{FF0000} \textbf{96.9± 3.6}} &
  \textbf{75.71± 6.4} &
  {\color[HTML]{FE0000} \textbf{81.70 ± 2.5}} &
  {\color[HTML]{FE0000} \textbf{66.12±5.4}} &
  {\color[HTML]{FE0000} \textbf{93.36± 2.3}} &
  {\color[HTML]{FF0000} \textbf{89.73±2.5}} &
  \textbf{80.3± 3.2} &
  {\color[HTML]{FF0000} \textbf{56.13 ± 3.0}} \\ \bottomrule
\end{tabular}}
\caption{Evaluation result on TU dataset using two different evaluation settings. The first one follows \cite{xu2018powerful}, and the second follows \cite{wijesinghe2021new}. We use † to denote the second setting. Top two results are highlighted by \textcolor{red}{\textbf{First}} and \textbf{Second}.}
\label{exp:tudataset}
\end{table*}

\textbf{Experiment setting.} For both graph property dataset and graph substructure counting dataset, GINE+~\cite{brossard2020graph} is used as the base encoder for SEK-GNN, and we also concatenate the hidden representation from a GIN model with SEK-GNN, followed by a linear transformation to get the final prediction. We call it \textit{SEK-GIN}. Similarly \textit{SEK-PPGN} leverages representation from PPGN and SEK-GNN where GIN is the base encoder for SEK-GNN.

For both datasets, we follow the standard data splitting method as in \citep{chen2020can,corso2020principal}. For graph property dataset, we use a learning rate of 8e-3, weight decay rate of 5e-7 for SEK-GIN, and $(1e-3, 0.0)$ for SEK-PPGN. For SEK-GIN, the model is trained for 350 epochs and for SEK-PPGN, the model is trained for 800 epochs. The batch sizes for SEK-GIN and SEK-PPGN are tuned over $\left\{64,128\right\}$. We use a hidden dim of 64 for both models. For SEK-GIN, the number of hops $k$ in GINE+ encoder is searched over $\left\{5,6\right\}$, and the number of layers is searched over $\left\{5,6,7\right\}$. For SEK-PPGN, we use a 1-layer GIN encoder for SEK-GNN, where the number of hops $k$ is searched over $\left\{3,5\right\}$. For SEK-GIN, we use \textit{ReduceLROnPlateau} learning rate scheduler with a patience of 10 and a reduction factor of 0.5. The pooling method used for jumping knowledge in SEK-GIN is attention-based pooling and \textit{summation} for SEK-PPGN. Finally, we use \textit{summation} as the $COMBINE$ function in SEK-PPGN and $geometric$ in SEK-GIN.

For the graph substructure counting dataset, we use a learning rate of 1e-3, and a weight decay rate of 1e-6 for SEK-GIN, and 1e-3 and 0.0 for SEK-PPGN. For SEK-GNN, the model is trained for 350 epochs, and for SEK-PPGN, the model is trained for 1000 epochs. The batch size for SEK-GIN and SEK-PPGN are tuned over $\left\{32,64\right\}$. We use a hidden dimension of 64 for both models. For SEK-GIN, the layer number of GINE+ encoder and the number of hops $k$ is searched over $\left\{(5,3),(5,5),(5,7),(6,3),(6,5),(6,7)\right\}$. For SEK-PPGN, we use a 1-layer GIN encoder for SEK-GNN, where the number of hops $k$ is searched over $\left\{3,5\right\}$. For SEK-GIN, we use \textit{ReduceLROnPlateau} learning rate scheduler with a patience of 15 and a reduction factor of 0.75. The pooling methods for jumping knowlwdge in SEK-GIN is attention-based pooling, and \textit{summation} for SEK-PPGN. Finally, we use \textit{summation} as $COMBINE$ function in SEK-PPGN and $geometric$ in SEK-GIN. The graph readout function is $summation$ in both SEK-PPGN and SEK-GIN.

\textbf{Results.} The evaluation metric for graph property dataset is $log_{10}{(MSE)}$, and for graph substructure counting dataset we use $MAE$.  Each model is trained three times and we report the average performance. As shown in Table \ref{exp:syn}, SEK-GIN or SEK-PPGN outperform the baseline methods on these two datasets for most tasks. Previously, we show that SEK-GNN can discriminate some non-isomorphic graphs where 3-WL fails, in these two synthetic datasets, we can see that a 1-layer 3-hop or 5-hop SEK-PPGN is able to uplift the expressive power of PPGN significantly, and sets new state-of-the-art performance in several tasks by a large margin. The most competitive baseline method to SEK-GNN is KP-GIN+, which encodes the internal substructures explicitly by leveraging the peripheral edges and induced subgraph, while SEK-GIN uses the substructure encoding function to encode the internal substructure implicitly.

\begin{table*}[t]
\centering
\begin{tabular}{@{}ccccccc|c@{}}
\toprule
Target & DTNN    & MPNN    & Deep LRP & PPGN          & Nested 1-2-3-GNN & KP-GIN+                                & SEK-GIN                                 \\ \midrule
$\mu$ &
  {\color[HTML]{000000} \textbf{0.244}} &
  0.358 &
  0.364 &
  {\color[HTML]{FF0000} \textbf{0.231}} &
  0.433 &
  0.367 &
  0.358 \\
$\alpha$      & 0.95    & 0.89    & 0.298    & 0.382         & 0.265            & \textbf{0.242}                         & {\color[HTML]{FF0000} \textbf{0.228}}   \\
$\varepsilon_{\text{HOMO}}$ &
  0.00388 &
  0.00541 &
  0.00254 &
  0.00276 &
  0.00279 &
  \textbf{0.00247} &
  {\color[HTML]{FF0000} \textbf{0.00106}} \\
$\varepsilon_{\text{LUMO}}$      & 0.00512 & 0.00623 & 0.00277  & 0.00287       & 0.00276          & \textbf{0.00238}                       & {\color[HTML]{FF0000} \textbf{0.00229}} \\
$\Delta \varepsilon$      & 0.0112  & 0.0066  & 0.00353  & 0.00406       & 0.0039           & \textbf{0.00345}                       & {\color[HTML]{FF0000} \textbf{0.00335}} \\
$\left\langle R^2\right\rangle$      & 17      & 28.5    & 19.3     & \textbf{16.7} & 20.1             & {\color[HTML]{FF0000} \textbf{16.49}}  & 16.91                                   \\
ZPVE      & 0.00172 & 0.00216 & 0.00055  & 0.00064       & \textbf{0.00015} & 0.00018                                & {\color[HTML]{FF0000} \textbf{0.00013}} \\
$U_0$      & 2.43    & 2.05    & 0.413    & 0.234         & 0.205            & \textbf{0.0728}                        & {\color[HTML]{FF0000} \textbf{0.0587}}  \\
$U$      & 2.43    & 2       & 0.413    & 0.234         & 0.2              & {\color[HTML]{FF0000} \textbf{0.0553}} & \textbf{0.0672}                         \\
$H$      & 2.43    & 2.02    & 0.413    & 0.229         & 0.249            & {\color[HTML]{FF0000} \textbf{0.0575}} & \textbf{0.073}                          \\
$G$     & 2.43    & 2.02    & 0.413    & 0.238         & 0.253            & {\color[HTML]{FF0000} \textbf{0.0526}} & \textbf{0.0592}                         \\
$C_v$     & 0.27    & 0.42    & 0.129    & 0.184       & {\color[HTML]{FF0000} \textbf{0.0811}}           & 0.0973                                 & \textbf{0.0924}  \\ \bottomrule
\end{tabular}
\caption{Evaluation result on QM9, Top two results are highlighted by \textcolor{red}{\textbf{First}} and \textbf{Second}.}
\label{exp:qm9}
\end{table*}

\subsection{Real-world datasets}
\subsubsection{Graph classification}
\textbf{Datasets.} We use eight datasets to evaluate the effectiveness of SEK-GNN, among which MUTAG, PROTEINS, ENZYMES, PTC-MR, BZR, and COX2 are bioinformatics datasets\citep{debnath1991structure,kriege2016valid,wale2008comparison,shervashidze2011weisfeiler,borgwardt2005shortest}, and IMDB-B and IMDB-M are social network datasets\cite{yanardag2015deep}.

\textbf{Baseline methods.} We compare against twelve baselines: (1) \textit{Graph kernel methods:} WL subtree kernel\cite{shervashidze2011weisfeiler}, RetGK\cite{zhang2018retgk}, GNTK\cite{du2019graph}, WWL \cite{togninalli2019wasserstein}, and FGW \cite{titouan2019optimal}. (2) \textit{GNN based methods:} DGCNN\cite{zhang2018end}, CapsGNN\cite{xinyi2018capsule}, GIN\cite{xu2018powerful}, GraphSAGE \cite{hamilton2017inductive}, GraphSNN\cite{wijesinghe2021new}, GIN-AK+\cite{zhao2021stars}, and KP-GNN \cite{feng2022powerful}. 

\textbf{Experiment setting.} We use 10-fold cross-validation, and the evaluation on validation set follows both settings and we report results for two settings\cite{xu2018powerful,wijesinghe2021new}. For the first setting\cite{xu2018powerful},  we use 9 folds for training and 1 fold for testing in each fold. After training, we average the test accuracy across all the folds. Then a single epoch with the best mean accuracy and the standard deviation is reported. For the second setting\cite{wijesinghe2021new}, we use 9 folds for training and 1 fold for testing in each fold but we directly report the mean best test results. We use SEK-GIN where the base GNN encoder for SEK-GNN is GINE+\cite{brossard2020graph}, we also use a GIN model and concatenate the graph representations followed by a linear transformation to get the final prediction logits. For all datasets, we search over: (1) the number of hops: $\left\{1,2,3,4\right\}$; (2) number of layers: $\left\{1,2,3,4\right\}$; (3) $COMBINE$ function: $\left\{sum,geometric\right\}$; (4)For JKNet, the pooling function is searched over $\left\{sum,concat\right\}$, and the graph readout function is summation. the learning rate is 8e-3 and weight decay rate is 1e-6, the hidden size is $max(int(120/K),40)$, where $K$ is the number of hops. 

\textbf{Result.} We report the model performance for both settings in Table \ref{exp:tudataset}. We can see that SEK-GIN achieves state-of-the-art performance for both settings. SEK-GIN outperforms GIN consistently across all the datasets, demonstrating that SEK-GNN can go beyond the expressive power of 1-WL test. Compared to other more powerful GNN models such as GIN-AK+, GraphSNN, and KP-GIN+, SEK-GNN can also achieve better performance or comparable performance. 

\subsubsection{Graph regression}
\textbf{Datasets.} we use QM9 dataset to verify graph regression tasks. QM9 contains 130K small molecules. The task here is to perform regression on twelve targets representing energetic, electronic, geometric, and
thermodynamic properties, based on the graph structure and node/edge features. 

\textbf{Baseline methods.} We compare against six baseline methods including DTNN and MPNN \cite{wu2018moleculenet}, PPGN\cite{maron2019provably}, Nested 1-2-3-GNN \cite{zhang2021nested} and KP-GNN \cite{feng2022powerful}. 

\textbf{Experiment setting.} We run three times and report the average results for every target. We use SEK-GIN where the base GNN encoder for SEK-GNN is GINE+, we also use a GIN model and concatenate the graph representations followed by a linear transformation to get the final prediction. We train a model separately for every target, we search over: (1) the number of hops: $\left\{5,6,7\right\}$, (2) number of layers: $\left\{4,6,8\right\}$, and (3) $COMBINE$ function: $\left\{sum,geometric\right\}$. For JKNet, the pooling function is attention-based aggregation, and the graph readout function is summation. The learning rate is 1e-3 and we don't use l2 weight decay and dropout, the hidden size is set to 128 for all targets. 

\textbf{Result.} As we can see in Table \ref{exp:qm9}, SEK-GIN also achieves state-of-the-art performance for many targets, where SEK-GIN is able to outperform PPGN and KP-GIN+ which are both upper bounded by 3-WL test. For other targets, SEK-GIN also achieves comparable performance with SOTA models, which demonstrates the effectiveness of the SEK 1-WL color refinement algorithm. 

\section{Conclusion}

In this paper, we establish connections between previous works on $K$-hop message-passing GNNs and 1-WL subgraph GNNs and find out that $K$-hop message-passing GNNs lack the ability to distinguish the internal substruture of subgraphs, we therefore propose to use \textit{substructure encoding function} as one possible mechanism to enhance the expressive power of $K$-hop message-passing GNNs. We further propose to utilize contextualized substructure information to uplift the expressiveness of $K$-hop message-passing GNNs. We then introduce SEK 1-WL test which is more expressive than $K$-hop 1-WL test and Subgraph 1-WL test. Finally, we provide a practical implementation, namely SEK-GNN which enjoys efficiency and parallelizability. Experiments on both synthetic datasets and real-world datasets demonstrate the effectiveness of SEK-GNN.

There are still other promising directions to explore. First, how to design other mechanisms to make $K$-hop message-passing GNNs to be able to distinguish the internal substructure of subgraphs with less computational cost compared with the proposed substructure encoding function. Second, how to establish connection between $K$-hop message-passing GNNs and other subgraph based GNN methods such as Equivariant Subgraph Aggregation Network (ESAN) \cite{bevilacqua2021equivariant} which is proven to be expressive for graph biconnectivity \cite{zhang2023rethinking}. This may provide a completely different view parallel to this work and guide us to design even more powerful $K$-hop message-passing GNN models. Finally, one can also exploit incorporating other structural encoding methods into $K$-hop message-passing GNNs to uplift the representative power. 

\section{Acknowledgements}

This work is partially supported by the NSF-Convergence Accelerator Track-D award \#2134901, by the National Institutes of Health (NIH) under Contract R01HL159805, by grants from Apple Inc., KDDI Research, Quris AI, IBT, and by generous gifts from Amazon, Microsoft Research, and Salesforce.

\newpage

\bibliographystyle{ACM-Reference-Format}
\balance
\bibliography{sample-base}

\appendix
\section{Appendix}
\subsection{PROOF OF THEOREM 1}
\label{proof theo1}
Our proof is inspired by the previous theoretical characterization on random regular graphs in \cite{li2020distance} and \cite{zhang2021nested}. We first outline our proof here: as $K$-hop message-passing GNNs also extract a node-induced subgraph implicitly, we can focus on the node-induced subgraph $G_u^{K}$ for any node $u \in \mathcal{V}$, then we show that $2K$ steps of randm walk is sufficiently large by restricting the random walk path inside $G_u^{K}$ and prune those paths outside $G_u^{K}$. Finally, we proof $2K$ steps of random walk is a sufficient condition to discriminate the internal substructure of $G_u^{K}$ by extending previous theoretical outcome from \cite{li2020distance}.

First, we define a path $P = (u_0,u_1,\cdots, u_d)$ is a tuple of nodes satisfying $\left\{u_{i-1},u_{i}\right\} \in \mathcal{E}, \forall i \in [d]$. We denote a path $P_{st}$ to be a path starting from node $s$ and end at node $t$, a self-return path can be denoted as $P_{uu}$ that start and end at node $u$. A path $P$ is said to be simple if it does not go through a node more than once, i.e., $u_i \neq u_j$ for $i \neq j$. The path $P$ induced by random walk is not necessarily a simple path. Next, we give a formal definition of \textit{edge configuration}. 

\textbf{Definition 6.} (\textbf{\textit{Edge configuration}}) The edge configuration between $\mathcal{N}_G^{k}(u)$ and $\mathcal{N}_G^{k+1}(u)$ for node $u \in \mathcal{V}$ is a list $C_{v, G}^k=\left(a_{v, G}^{1, k}, a_{v, G}^{2, k}, \ldots\right)$, where $a_{v, G}^{i, k}$ denotes the number of nodes in $\mathcal{N}_G^{k+1}(u)$ of which each has exactly $i$ edges from $\mathcal{N}_G^{k}(u)$.

For two nodes $v_1$ and $v_2$, in $G_1$ and $G_2$, the two edge configurations $C_{v_1, G_1}^k$ equals to $C_{v_2, G_2}^k$ if and only if $C_{v_1, G_1}^k$ are component-wise equal to $C_{v_2, G_2}^k$. Next, we propose the first lemma. 

\begin{lemma}
\label{lemma1}
For two graphs $G_1$ and $G_2$ that are uniformly independently sampled from all $n$-node $r$-regular graphs, where $3 \leq r<\sqrt{2 \log n}$, we pick any two nodes, each from one graph, denoted by $v_1$ and $v_2$ respectively. Then, there is at least one $i$ that is greater than $\frac{1}{2} \frac{\log n}{\log (r-1-\epsilon)}$ and less than $\left(\frac{1}{2}+\epsilon\right) \frac{\log n}{\log (r-1-\epsilon)}$ with probability $1-o\left(n^{-1}\right)$ such that $C_{v_1, G_1}^{i} \neq C_{v_2, G_2}^{i}$. Moreover, with at least the same probability, for all $i \in\left(\frac{1}{2} \frac{\log n}{\log (r-1-\epsilon)},\left(\frac{2}{3}-\epsilon\right) \frac{\log n}{\log (r-1)}\right)$, the number of edges between $\mathcal{N}_{G_{j}}^{i}(v_j)$ and $\mathcal{N}_{G_{j}}^{i+1}(v_j)$ are at least $(r-1-\epsilon)\left|\mathcal{N}_{G_{j}}^{i}(v_j)\right|$ for $j \in\{1,2\}$.

\end{lemma}

\textit{Proof.} This lemma can be obtained by following step 1-3 in the proof of Theorem 3.3 in \cite{li2020distance}. 

Using lemma \ref{lemma1}, we now set $K=\left\lceil\left(\frac{1}{2}+\epsilon\right) \frac{\log n}{\log (r-1-\epsilon)}+1\right\rceil$, then for the two induced subgraphs $G_{v_1}^K$ and $G_{v_2}^K$, we can show that the with probability $1-o\left(n^{-1}\right)$, the total number of self-return paths $\left|P_{{v_1}{v_1}}\right| \neq \left|P_{{v_2}{v_2}}\right|$ where the length of path $P_{{v_1}{v_1}}$ and $P_{{v_2}{v_2}}$ are at most $2K$.

\begin{lemma}
\label{lemma2}
For two graphs $G_1$ and $G_2$ that are uniformly independently sampled from all $n$-node $r$-regular graphs, where $3 \leq r<\sqrt{2 \log n}$, we pick any two nodes, each from one graph, denoted by $v_1$ and $v_2$ respectively. For the two node-induced subgraphs $G_{v_1}^K$ and $G_{v_2}^K$ with $K=\left\lceil\left(\frac{1}{2}+\epsilon\right) \frac{\log n}{\log (r-1-\epsilon)}+1\right\rceil$, the total number of self-return paths $P_{{v_1}{v_1}}$ will not be equal to that of $P_{{v_2}{v_2}}$, i.e., $\left|P_{{v_1}{v_1}}\right| \neq \left|P_{{v_2}{v_2}}\right|$,with probability roughly $1-o\left(n^{-1}\right)$.
\end{lemma}

\textit{Proof.} Using lemma \ref{lemma1}, we know there exists some $k<K$, with probability $1-o\left(n^{-1}\right)$ such that $C_{v_1, G_1}^{k} \neq C_{v_2, G_2}^{k}$, then if we focus on the paths $P$ reaching to $\mathcal{N}_{G_{1}}^{k+1}(v_1)$ and  $\mathcal{N}_{G_{2}}^{k+1}(v_2)$ but not further, then the total number of paths $\left|P_{{v_1}{v_1}}\right|$ and $\left|P_{{v_2}{v_2}}\right|$ is likely to be different, due to the fact that $C_{v_1, G_1}^{k} \neq C_{v_2, G_2}^{k}$, with probability $1-o\left(n^{-1}\right)$. However, we notice that there is some chance that $\sum_{i=1}^r i \times a_{v_1, G_1}^{i, k} = \sum_{i=1}^r i \times a_{v_2, G_2}^{i, k}$, i.e., although $C_{v_1, G_1}^{k} \neq C_{v_2, G_2}^{k}$ component-wise, the total number of edges going from $\mathcal{N}_{G_{1}}^{k+1}(v_1)$ to $\mathcal{N}_{G_{1}}^{k}(v_1)$ and $\mathcal{N}_{G_{2}}^{k+1}(v_2)$ to $\mathcal{N}_{G_{2}}^{k}(v_2)$ is the same. Hence, we say that $\left|P_{{v_1}{v_1}}\right| \neq \left|P_{{v_2}{v_2}}\right|$ with probability roughly $1-o\left(n^{-1}\right)$.

Lemma \ref{lemma2} also suggests that $2K$ steps is sufficient to lead to $\left|P_{v_1 v_1}\right| \neq \left|P_{v_2 v_2}\right|$ as $k$ is strictly less than $K$. Now we show that $\left|P_{v_1 v_1}\right| \neq \left|P_{v_2 v_2}\right|$ is a sufficient condition for a self-return probability vector of $2K$ steps $\left\{H_{uu}^t\right\}_{t=1}^{2K}$ to differentiate the internal substructure of $G_{v_1}^K$ and $G_{v_2}^K$.

First, for some $k<K$ leading to $C_{v_1, G_1}^{k} \neq C_{v_2, G_2}^{k}$ with probability $1-o\left(n^{-1}\right)$, the total sample space (total number of paths) for $v_1$ and $v_2$ as $G_1$ and $G_2$ are both $n$-node $r$-regular graphs, and the total sample space is independent of the subgraph $G_{v_1}^K$ and $G_{v_2}^K$. 

Next, we consider three types of paths:
i) The paths touching $\left\{\mathcal{N}_{G_{1}}^{i}(v_1)\right\}_{i=1}^{k}$ and $\left\{\mathcal{N}_{G_{2}}^{i}(v_2)\right\}_{i=1}^{k}$ but not further.
ii) The paths touching $\left\{\mathcal{N}_{G_{1}}^{k+1}(v_1)\right\}$ and $\left\{\mathcal{N}_{G_{2}}^{k+1}(v_2)\right\}$ but not further.
iii) The paths touching $\left\{\mathcal{N}_{G_{1}}^{i}(v_1)\right\}_{i=k+2}^{K}$ and $\left\{\mathcal{N}_{G_{2}}^{i}(v_2)\right\}_{i=k+2}^{K}$ but not further, if $k+2<K$.

For type 1 paths, the total number of self-return paths to $v_1$ and $v_2$ remains the same due to the symmetry of the subgraph. For type 2 paths, using Lemma \ref{lemma2}, we can see that the total number of paths for $v_1$ and $v_2$ is different with probability roughly $1-o\left(n^{-1}\right)$. For type 3 paths, the total number of self-return paths is also dependent on $C_{v_1, G_1}^{k}$ and $C_{v_2, G_2}^{k}$. 

Finally, there are paths that touch $\left\{\mathcal{N}_{G_{1}}^{i}(v_1)\right\}_{i=K+1}^{2K}$ and $\left\{\mathcal{N}_{G_{2}}^{i}(v_2)\right\}_{i=K+1}^{2K}$; however, these paths don't get back to $v_1$ and $v_2$, hence they don't count as valid paths. 

Based on the above observations, we can see that the self-return probability vector starts to make a difference at timestep $2(k+1)$ and onwards. This concludes the proof that a self-return probability vector of length $2K$ is sufficient for distinguishing the internal substructure of two induced subgraphs.

Finally we make some notes on the cases when $C_{v_1, G_1}^{k} \neq C_{v_2, G_2}^{k}$, but $\sum_{i=1}^r i \times a_{v_1, G_1}^{i, k} = \sum_{i=1}^r i \times a_{v_2, G_2}^{i, k}$, which will make the self-return probability vector indistinguishable. We call it \textit{collision rate}, and let it be a scalar value $p$, however as we inject contextualized substructures, to make it indistinguishable for $G_{v_1}^K$ and $G_{v_2}^K$, it is required that all the nodes $i \in G_{v_1}^K$ and $j \in G_{v_2}^K$ collides and $\left|G_{v_1}^K\right| = \left|G_{v_2}^K\right|$. Therefore, the collision rate now reduces to $p^{\left|G_{v_1}^K\right|}$, which reduces exponentially as $\left|G_{v_1}^K\right|$, the total number of nodes in $G_{v_1}^K$ grows. The collision rate becomes negligible as $\left|G_{v_1}^K\right|$ increases. This demonstrates the benefits to incorporate contextualized substructure information. 

\subsection{Ablation Study}
\label{Ablation_Study}

\begin{table}[]
\centering
\scalebox{0.63}{
\begin{tabular}{@{}lccccccc@{}}
\toprule
\textbf{}           & \multicolumn{3}{c}{\textbf{Substructure counting}}  & \multicolumn{4}{c}{\textbf{Graph property regression}}                \\ \cmidrule(l){2-8} 
\multicolumn{1}{c}{\textbf{Method}} & \textbf{Tri.} & \textbf{Tailed Tri.} & \textbf{Star} & \textbf{4-cycle} & \textbf{Connect.} & \textbf{Diameter} & \textbf{Radius} \\ \midrule
SEK-GIN(\#steps=0)  & 0.1529          & 0.0891          & 0.0011          & 0.0948          & -2.215          & -2.647          & -4.823          \\
SEK-GIN(\#steps=8)  & 0.0043          & \textbf{0.0059} & 0.0008          & 0.0039          & -2.439          & -3.078          & -4.915          \\
SEK-GIN(\#steps=16) & 0.0034          & 0.0061          & \textbf{0.0007} & \textbf{0.0037} & -2.555          & -3.268          & -4.902          \\
SEK-GIN(\#steps=24) & 0.0034          & 0.0068          & 0.0007          & 0.0043          & \textbf{-2.765} & -3.423          & -5.005          \\
SEK-GIN(\#steps=32) & \textbf{0.0033} & 0.0066          & 0.0008          & 0.0043          & -2.663          & \textbf{-3.591} & \textbf{-5.103} \\ \bottomrule
\end{tabular}}
\caption{Ablation study on synthetic datasets using SEK-GIN}
\label{exp:ab_study1}
\end{table}

We further perform a detailed ablation study on our proposed framework on both synthetic datasets and real-world datasets to demonstrate the effectiveness of our proposed method. One core contribution in our work is the proposal of a general notion of substructure encoding function and one instantiation of it which achieves parallelizability and theoretical guarantee to encode the substructural information. Hence, we perform ablation study to evaluate the effect of various step sizes of random-walk, in addition to that without injecting any substructure information(\#steps=0). 

For synthetic datasets, we fix the SEK-GIN to be 6-layer 5-hop with summation as combine function and graph readout function. The model is trained for 400 epochs and three times for each target and we report the average result. For TU dataset, the model is fixed to be a 2-layer 3-hop SEK-GIN, where the combine function is summation and the graph readout function is also summation. All models are trained for 350 epochs. We follow the second evaluation setting\cite{wijesinghe2021new} to report the experiment result across 10 folds. The result is illustrated in Table \ref{exp:ab_study1} and Table \ref{exp:ab_study2}. As we can see, the performance gains for SEK-GIN are significant when the contextualized structural information is injected, which demonstrates the effectiveness of our proposed method. 


\begin{table}[]
\centering
\scalebox{0.72}{\begin{tabular}{ccccc}
\hline
Method              & IMDB-BINARY   & PROTEINS      & BZR           & MUTAG         \\ \hline
SEK-GIN(\#steps=0)  & 0.781 ± 0.037 & 0.790 ± 0.030 & 0.906 ± 0.021 & 0.936 ± 0.052 \\
SEK-GIN(\#steps=8)  & 0.791 ± 0.036 & 0.796 ± 0.035 & 0.911 ± 0.030 & 0.952 ± 0.039 \\
SEK-GIN(\#steps=16) & 0.792 ± 0.043 & 0.803 ± 0.034 & 0.911 ± 0.019 & 0.952 ± 0.039 \\
SEK-GIN(\#steps=24) & 0.785 ± 0.041 & 0.803 ± 0.029 & 0.912 ± 0.019 & 0.952 ± 0.029 \\
SEK-GIN(\#steps=32) & 0.787 ± 0.034 & 0.808 ± 0.027 & 0.921 ± 0.022 & 0.958 ± 0.033 \\ \hline
\end{tabular}}
\caption{Ablation study on TU datasets using SEK-GIN}
\label{exp:ab_study2}
\end{table}

\subsection{Definition of distance-regular graph and intersection array}
\label{definition}
In this section, we give a formal definition of distance-regular graph and intersection array.

\textbf{Definition 7.} (\textbf{\textit{Distance-regular graph}}) Given a graph $G=(\mathcal{V}, \mathcal{E})$ and let $D(G):=\max _{u, v \in \mathcal{V}} \operatorname{dis}_G(u, v)$ be the diameter of $G$. We say $G$ is distance-regular if for all $i, j \in[D(G)]$ and all nodes $u,v,w,x \in \mathcal{V}$ with $dis_G(u,v)= dis_G(w,x)$, we have $\left|\mathcal{N}_G^i(u) \cap \mathcal{N}_G^j(v)\right|=\left|\mathcal{N}_G^i(w) \cap \mathcal{N}_G^j(x)\right|$.

\textbf{Definition 8.} (\textbf{\textit{Intersection array}}) The intersection array of a distance-regular graph $G$ is denoted as: 

$\iota(G)=\left\{b_0, \cdots, b_{D(G)-1} ; c_1, \cdots, c_{D(G)}\right\}$, where $b_i=\left|\mathcal{N}_G(u) \cap \mathcal{N}_G^{i+1}(v)\right|$ and $c_i=\left|\mathcal{N}_G(u) \cap \mathcal{N}_G^{i-1}(v)\right|$ with $\operatorname{dis}_G(u, v)=i$.

\subsection{Datasets Statistics}
\label{appendix:dataset_stats}
\begin{table}[H]
\scalebox{0.8}{
\begin{tabular}{@{}ccccc@{}}
\toprule
Dataset               & \# Tasks & \# Graphs      & Ave. \# Nodes & Ave. \# Edges \\ \midrule
Graph property        & 3       & 5120/640/1280  & 19.5          & 101.1         \\
Substructure counting & 4       & 1500/1000/2500 & 18.8          & 62.6          \\ \midrule
MUTAG                 & 2       & 188            & 17.93         & 19.79         \\
PTC\_MR               & 2       & 344            & 14.29         & 14.69         \\
PROTEINS              & 2       & 1113           & 39.06         & 72.82         \\
ENZYMES               & 6       & 600            & 32.63         & 62.14         \\
BZR                   & 2       & 405            & 35.75         & 38.36         \\
COX2                  & 2       & 467            & 41.22         & 43.45         \\
IMDB-B                & 2       & 1000           & 19.77         & 96.53         \\
IMDB-M                & 3       & 1500           & 13            & 65.94         \\ \midrule
QM9                   & 12      & 129433         & 18            & 18.6          \\ \bottomrule
\end{tabular}}
\caption{Datasets statistics}
\end{table}

\end{document}